\documentclass{article}

    \PassOptionsToPackage{numbers, compress}{natbib}

\usepackage[preprint]{neurips2023}

\usepackage[utf8]{inputenc} 
\usepackage[T1]{fontenc}    
\usepackage{hyperref}       
\usepackage{url}            
\usepackage{booktabs}       
\usepackage{amsfonts}       
\usepackage{nicefrac}       
\usepackage{microtype}      
\usepackage{xcolor}         

\usepackage[ruled,linesnumbered]{algorithm2e}
\usepackage{bbm}
\usepackage{algpseudocode}
\usepackage{multirow}
\usepackage{colortbl}  
\usepackage{xcolor}
\usepackage{array}   
\usepackage{soul}
\usepackage{amsmath}
\usepackage{float}
\usepackage{graphicx}
\usepackage{subfig}

\newtheorem{theorem}{Theorem}[section]

\newtheorem{assumption}[theorem]{Assumption}

\title{Unleash Model Potential: Bootstrapped Meta Self-supervised Learning}

\author{%
  Jingyao Wang, Zeen Song, Wenwen Qiang*, Changwen Zheng \\
  Institute of Software Chinese Academy of Sciences\\
  \texttt{\{wangjingyao2023, songzeen, qiangwenwen, changwen\}@iscas.ac.cn}
}

\begin{document}

\maketitle

\begin{abstract}
      The long-term goal of machine learning is to learn general visual representations from a small amount of data without supervision, mimicking three advantages of human cognition: i) no need for labels, ii) robustness to data scarcity, and iii) learning from experience. Self-supervised learning and meta-learning are two promising techniques to achieve this goal, but they both only partially capture the advantages and fail to address all the problems. Self-supervised learning struggles to overcome the drawbacks of data scarcity, while ignoring prior knowledge that can facilitate learning and generalization. Meta-learning relies on supervised information and suffers from a bottleneck of insufficient learning. To address these issues, we propose a novel Bootstrapped Meta Self-Supervised Learning (BMSSL) framework that aims to simulate the human learning process. We first analyze the close relationship between meta-learning and self-supervised learning. Based on this insight, we reconstruct tasks to leverage the strengths of both paradigms, achieving advantages i and ii. Moreover, we employ a bi-level optimization framework that alternates between solving specific tasks with a learned ability (first level) and improving this ability (second level), attaining advantage iii. To fully harness its power, we introduce a bootstrapped target based on meta-gradient to make the model its own teacher. We validate the effectiveness of our approach with comprehensive theoretical and empirical study.
\end{abstract}


\section{Introduction}
\label{Introduction}

Humans are able to understand the world with three advantages \cite{mcardle2014human, carroll1993human}: i) no need for supervised information; ii) only a small number of samples is required for recognizing a classification task; and iii) learning based on the existing prior knowledge of the world. Correspondingly, the ultimate aim of machine learning is to leverage prior knowledge and learn representations that can transfer across different tasks without requiring any supervision, even when the data is scarce.

\begin{figure}
\centering
\begin{minipage}[t]{0.48\textwidth}
\centering
\includegraphics[width=\textwidth]{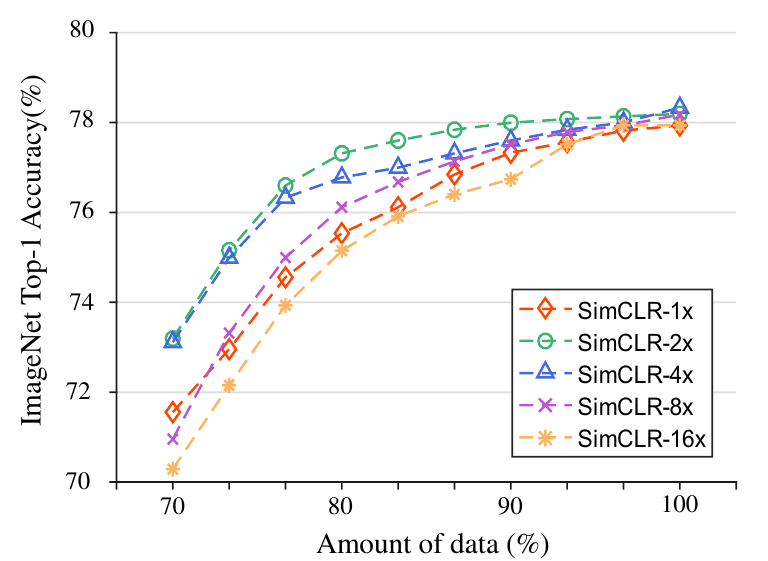}
\caption{Accuracies with linear evaluation for different scales of data and augmentation. "SimCLR-$N$x" means that multiple data enhancements are used randomly to expand the data to $N$ times.}
\label{fig:intro_fig}
\end{minipage}
\hspace{.15in}
\begin{minipage}[t]{0.48\textwidth}
\centering
\includegraphics[width=\textwidth]{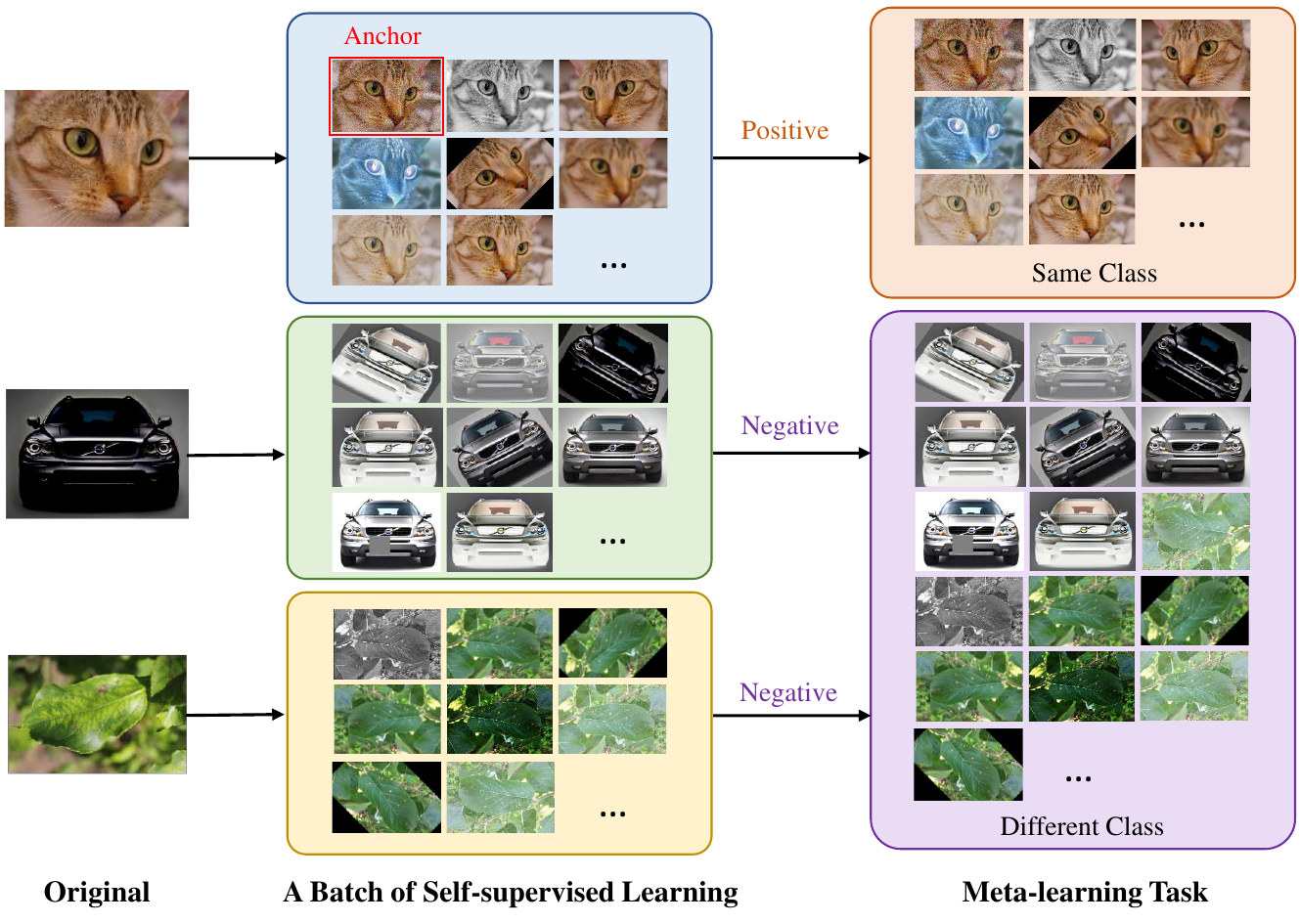}
\caption{Restructuring SSL data into meta-Learning tasks. Views augmented from the same "cat" with the \textbf{Anchor} are positive (same class), while the views of other samples (the “car” and the “leaf”) are negative (different class).}
\label{fig:intro_MetaSSL-BMG}
\end{minipage}
\end{figure}

Self-supervised learning (SSL) is a promising approach to achieve this goal, as it can learn general representations without supervision and generalize to downstream tasks \cite{krishnan2022self, schiappa2022self, shurrab2022self, baevski2022data2vec}. SSL applies various data augmentations \cite{shorten2019survey, yang2022image} to generate different views of the same image and encourages them to have similar embeddings while being dissimilar from views obtained by other images. SSL has been considered as a close approximation to human learning in machine learning \cite{albelwi2022survey, liu2021self, jaiswal2020survey}. However, we argue that it only partially captures the first advantage of human learning and fails to address the other two. Specifically, we show that data augmentation cannot fully compensate for the lack of data diversity and may even harm the performance when overused (see Figure \ref{fig:intro_fig}). Moreover, we point out that SSL relies on a single data-based fixed prior, e.g. data distribution need to satisfy uniform distribution, which may introduce bias when data is scarce and limit the adaptability to new scenarios. Therefore, SSL still faces significant challenges in overcoming the low-data barrier and incorporating flexible prior knowledge as humans do.

Meta-learning is another promising approach for this goal that aims to overcome the low-data barrier by learning to quickly adapt to new tasks with limited data \cite{huisman2021survey, chen2021meta, hospedales2021meta, sun2019meta, vanschoren2018meta}. Meta-learning mimics human learning by using a double loop structure: the inner loop optimizes a task-specific model based on the current learning ability, and the outer loop updates this ability based on the feedback from multiple tasks. However, meta-learning still relies on supervision to update the ability, which violates the first advantage of human learning. Some recent works attempt to use self-supervised pseudo-labels for meta-learning to avoid supervision \cite{amac2022masksplit, fang2021ssml, bansal2020self}, but this strategy is computationally expensive and may not generalize well to different tasks. Furthermore, meta-learning suffers from two limitations: i) inner loop: the task-specific model $f$ is only updated for $L$ steps, which may not capture the optimal state for the task; ii) outer loop: the learning ability update is based on the same objective and geometry as $f$, which may propagate the errors of $f$ and compromise the final results. Therefore, meta-learning still has a large gap from achieving human-like learning capabilities.

In this work, we reveal a surprising connection between self-supervised learning and meta-learning: they can be unified by viewing the batch of classes sampled in meta-learning as the augmented views of samples generated in self-supervised learning. Based on this insight, we propose a novel Bootstrapped Meta Self-Supervised Learning framework (BMSSL) that aims to emulate the three advantages of human learning. To achieve the first and second advantages, we present a simple and general method to restructure tasks that can leverage both self-supervised learning and meta-learning, overcoming the limitations of labels and data volume. Figure \ref{fig:intro_MetaSSL-BMG} briefly illustrates our idea. To achieve the third advantage, we use a bi-level meta-learning structure with gradient-based optimization to update the initial parameters based on feedback from multiple tasks. To address the existing bottlenecks of meta-learning, we further introduce a bootstrapped target based on meta-gradient to allow the model to learn from itself. Our contributions are as follows: 

\begin{itemize}
    \item We discuss the close relationship between meta-learning and self-supervised learning, and experimentally verify the feasibility of our inductive ideas. 
    \item We propose a novel Bootstrapped Meta Self-Supervised framework (BMSSL) to simulate the learning process of human.
    \item We conduct theoretical analysis and empirical study on our proposed framework to verify its effectiveness.
\end{itemize}


\section{Related Work}
\label{Related Work}

\textbf{Self-supervised learning}.
Self-supervised learning (SSL) enables learning of general visual representations by imposing an additional constraint between different views of raw input data without accessing any annotated data \cite{ elnaggar2021prottrans, li2021cutpaste, bachman2019learning, foster2020improving, tian2020makes}. Recently, mainstream approaches rely on constructing positive and negative viewpoints for examples through augmentations \cite{ren2022simple, jaiswal2020survey, ren2022simple} to learn pretext tasks and generalize to downstream tasks: positive samples are typically augmented views of the same reference instance, while negative examples are defined as any view from a different instance. The learning process is based on the pioneering NCE method \cite{oord2018representation}, which uses contrastive loss \cite{zhu2021graph, zimmermann2021contrastive} to enforce discrimination between positive and negative viewpoints of each instance, leading to the learning of useful semantics. The learned model provides visually discriminative representations uniformly scattered in the feature space. However, these methods are difficult to generalize when data is scarce, and learn based on only one stage of data instead of quickly adapting to new tasks from experience with multiple learning processes like humans.

\textbf{Meta-learning}.
Meta-learning aims to learn a model that can quickly adapt to new tasks with limited data and generalize to unseen examples. The meta-learning methods can be divided into two categories: i) learn the optimal initialization to adapt to new tasks quickly \cite{finn2017model, nichol2018reptile, kao2021maml}; ii) learn a shared embedding space and amortizing inference \cite{vinyals2016matching, snell2017prototypical, sung2018learning, zhu2022convolutional}. Recently, meta-learning has achieved superior performance in various applications, such as few-shot classification \cite{zhang2021shallow, ren2018meta}, reinforcement learning \cite{pike2022reinforcement, zhao2022offline}, and hyperparameter optimization \cite{zhang2018fine}. These models cleverly design tasks that rely on a few labeled samples to learn a general visual representation unit, but often fails to provide reliable uncertainty estimates when only a few meta-training tasks or on supervision is provided \cite{luo2022meta, huisman2021survey}. Some methods adopt a patchwork approach to solve this problem: using unsupervised model to construct pseudo-labels, then use them as supervision for meta-learning. However, although this approach can learning representations from limited data without human priors, it leads to huge computational resources while be difficult to guarantee accuracy \cite{park2020meta}. Meanwhile, due to the myopia and curvature limitations of meta-learning \cite{flennerhag2021bootstrapped}, it still cannot meet expectations about simulating human learning.

\section{The Relationship between Meta-Learning and Self-supervised Learning}
\label{Study of the Relationship}
In this section, we study the close relationship between self-supervised learning and meta-learning. We first review the learning paradigms of both frameworks and highlight their similarities from three perspectives. Based on these insights, we propose a simple meta-learning-based approach to reconstruct self-supervised tasks and evaluate its effectiveness experimentally.

\begin{figure}
\begin{center}
\includegraphics[width=\textwidth]{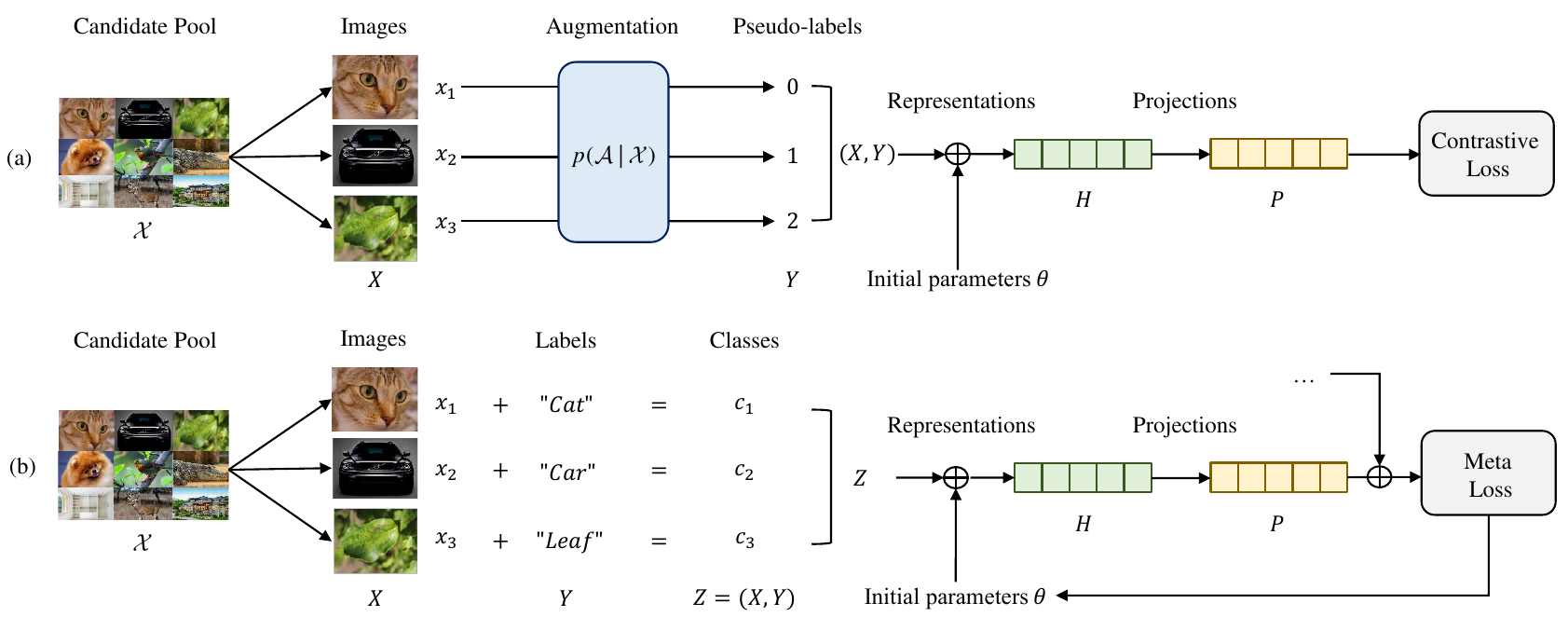}
\end{center}
   \caption{(a) Training procedure of contrastive learning. Augmented views from images $X$ are generated by applying random transformations to the same input batch, obtain corresponding pseudo-labels $Y$. $H$ and $P$ are vectors through a backbone for representations and a projector learned through contrastive prediction tasks. (b) Training procedure of meta-learning. Its input is labeled data $Z$, and learns a task-specific model based on a learnable initialization parameter $\theta$. The subsequent processing of each task is similar to self-supervised learning, but $\theta$ is updated based on the cumulative gradient of all tasks.}
\label{fig:relationship}
\end{figure}

The training procedures of meta-learning and self-supervised learning are shown in Figure \ref{fig:relationship}. For self-supervised learning: i) sample $ \{ x  \}_{i}^{N}$ from the distribution of the raw data space $\mathcal{X} $; ii) apply multiple data augmentations to $ \{ x  \}_{i}^{N}$, e.g., random scaling, rotation, and cropping, obtaining $\mathcal{A} = \{ a_{i}^{x_{1}}, \dots, a_{i}^{x_{N}}  \}_{i=1}^{M}$; iii) learn a general visual representation based on minimizing self-supervised learning loss, e.g., contrastive loss; iv) use pre-trained models to extract features and verify them in downstream tasks. For meta-learning: i) sample ${x}_{i}^{N}$ from the distribution of the raw data space $X$; ii) construct tasks $ \{ T  \}_{i}^{K}$ by partitioning ${x}_{i}^{N}$; iii) find the best initial parameters based on meta loss (outer loop): minimizes the cumulative gradient loss for all of the task-specific model (inner loop); iv) use the trained model for fast adaptation to new tasks. Conceptually, we find that the training process of self-supervised learning is similar to meta-learning and includes several similarities:
\begin{itemize}
    \item Both aim to learn generalizable representations and quickly adapt to new tasks: self-supervised learning aims to discriminate between unseen images; meta learning aims to discriminate between unseen tasks. 
    \item Both learn a fixed amount of information that can be transferred to new tasks: self-supervised learning leverages the similarity and the dissimilarity among multiple views of the samples; meta-learning leverages the similarity of instances within each task. 
    \item Both use batches as units of data processing: self-supervised learning treats all views generated by a single augmentation as a batch; meta learning treats each task which consists of $K$ $n$-way-$m$-shot tasks as a batch. 
\end{itemize}

Inspired by this, we propose a general paradigm to unify self-supervised learning and meta-learning. The idea is to transform self-supervised learning into a task distribution that is suitable for meta-learning optimization and learn from it. The paradigm consists of the following steps: i) randomly sample a batch of $N$ input images $ \left \{ x_{i} \right \} _{i=1}^{N} \in \mathcal{X} $ from the candidate pool; ii) divide $\mathcal{X}$ into $K$ blocks, each block contains $N/K$ images; iii) apply multiple data augmentations on $x \in \mathcal{X}$ of each blocks, obtaining $a \in \mathcal{A}$; iv) create an $N$-way classification problem for each block: all data generated from the same $x_{i}$ is regarded as a category, with $z=\{a,y\}\in\mathcal{Z}$ as the data and $y\in \mathcal{Y}$ as the pseudo-labels; v) integrate tasks corresponding to $K$ blocks in a batch, obtaining the task distribution $p(\mathcal{T})$; vi) update the task-specific model $w$ for task $T_{i}$ based on the meta-initialized parameters $\theta \in \zeta $ by calculating the loss $l(\cdot)$ (\emph{inner loop}); vii) use the meta-objective function $\mathcal{L} (\cdot,\theta)$ to update $\theta$, which is calculate through $\arg\min  \frac{1}{K} \sum_{i=1}^{K}l(w_{i}) $  (\emph{outer loop}). Before further extending this paradigm, we state the assumptions needed for this paradigm. Theoretical analysis details can be found in Section \ref{Theoretical Analysis} and Appendix A.


\begin{assumption}\label{assum:Task}
    We assume that $\mathcal{Z}$ is a Polish space (i.e., complete, separable, and metrizable), and for any $i$, $p(T_{i})$ is a non-atomic probability distribution on $(\mathcal{Z},\mathcal{L})$, where $\mathcal{L}(\cdot,\theta)$ is a Borel $\sigma$-algebra on $\mathcal{Z}$.
\end{assumption}

\begin{assumption}\label{assum:para}
    The ground truth parameter $\theta^{*}$ is independent of $\mathcal{X}$ and satisfies $\mathrm {Cov}[\theta^*]=(R^2/d)\mathrm {} I_d  $, where $R$ is a constant and $d$ is the dimension of the model parameter.
\end{assumption}

\begin{assumption}\label{assum:OF}
    For any $z\in\mathcal{Z}$, the function $\mathcal{L} (\cdot,\theta)$ is twice continuously differentiable, and satisfies the following properties for any $z\in\mathcal{Z}$, $w_i,w_j\in \mathbb{R}^d $:
    \begin{itemize}
    \item The function $\mathcal{L}(\cdot,\theta)$ is $K$-bounded over $\mathcal{W}$ with the gradient norm uniformly bounded by $G$, i.e., $\left \| \bigtriangledown l(z,w_i) \right \| \le G$;
    \item The function $\mathcal{L}(\cdot,\theta)$ is L-smooth over $\mathbb{R}^d$, i.e., $\left \| \bigtriangledown l(z,w_i)-\bigtriangledown l(z,w_j) \right \|-L(w_i-w_j)  \le  0$;
    \item The function $\mathcal{L}(\cdot,\theta)$ is $\mu $-strongly convex, i.e., $\left \| \bigtriangledown l(z,w_i)-\bigtriangledown l(z,w_j) \right \|-\mu(w_i-w_j)  \ge  0$.
    \end{itemize}
\end{assumption}

To evaluate the feasibility of the proposed general paradigm, we compare the performance of typical self-supervised frameworks, SimCLR \cite{chen2020simple}, Barlow Twins \cite{zbontar2021barlow}, and MoCo \cite{he2020momentum}, in learning representations under different settings. We measure the top-1 accuracies on the ImageNet1K \cite{wang2022vim} and CIFAR-10 \cite{krizhevsky2009learning} datasets without imposing any data restrictions, where pure self-supervised frameworks can perform well. We expect that the variants of the model based on the proposed paradigm can benefit from it or at least not degrade in performance in such environments. The results in Tables \ref{tab:relation1} and \ref{tab:relation2} confirm this expectation. We observe that the self-supervised learning representations based on the proposed paradigm can achieve comparable or even better performance than the original frameworks on both datasets. Moreover, since the improvement is limited in this setting, we conduct a more comprehensive assessment from multiple perspectives in Section \ref{Experiments}.

\begin{minipage}{\textwidth}
\begin{minipage}[t]{0.48\textwidth}
\makeatletter\def\@captype{table}
\caption{Accuracies(\%) on ImageNet1K on SSL baselines ("-o" means using the meta-learning-based setting, while "-x" means not).}
\resizebox{\linewidth}{!}{
\begin{tabular}{lcc}
    \toprule
    \textbf{Method} & \textbf{Backbone}  & \textbf{Top-1 Acc} \\
    \midrule
    SimCLR-x        & ResNet-50    & 64.561  \\
    SimCLR-o        & ResNet-50    & 65.156  \\
    Barlow Twins-x  & ResNet-50    & 66.561  \\
    Barlow Twins-o  & ResNet-50    & 66.952  \\
    Moco-x          & ResNet-50    & 59.382  \\
    Moco-o          & ResNet-50    & 59.156  \\
    \bottomrule
\end{tabular}}
\label{tab:relation1}
\end{minipage}
\hspace{.1in}
\begin{minipage}[t]{0.48\textwidth}
\makeatletter\def\@captype{table}
\caption{Accuracies(\%) on CIFAR-10 on SSL baselines ("-o" means using the meta-learning-based setting, while "-x" means not).}
\resizebox{\linewidth}{!}{
\begin{tabular}{lcc}
    \toprule
    \textbf{Method} & \textbf{Backbone}  & \textbf{Top-1 Acc}\\
    \midrule
    SimCLR-x        & ResNet-18    & 89.416  \\
    SimCLR-o        & ResNet-18    & 91.516  \\
    Barlow Twins-x  & ResNet-18    & 92.513  \\
    Barlow Twins-o  & ResNet-18    & 92.789  \\
    Moco-x          & ResNet-18    & 82.465  \\
    Moco-o          & ResNet-18    & 83.165  \\
    \bottomrule
\end{tabular}}
\label{tab:relation2}
\end{minipage}
\end{minipage}


\section{A Bootstrapped Meta Self-Supervised Learning Framework}
\label{Algorithm}
In this section, we introduce the proposed Bootstrapped Meta Self-Supervised Learning framework (BMSSL), which is inspired by the human cognitive model. Our key idea includes: i) construct few-shot classification tasks using a multi-view queue generated by data augmentation in self-supervised learning (Section \ref{Online Task Construction}); ii) use a gradient-based two-level optimization structure to learn a universal visual representation (Section \ref {Bootstrapped Meta Self-supervised Learning}): the inner loop use contrastive learning to learn specific task (first optimization); the outer loop find the optimal initial parameters for inner loop with bootstrapped meta gradient through minimizing the distance to the self-bootstrapped objective, which allows the model to become its own teacher (second optimization). The pseudocode of our framework is provided in Appendix A. It is worth noting that BMSSL is designed based on the observation in Section \ref{Study of the Relationship}, thus adopting consistent notation and settings.

\subsection{Online Task Construction}
\label{Online Task Construction}

We now describe how to re-construct self-supervised task based on meta-learning for BMSSL. Firstly, we randomly select $N$ unlabelled data $x \in \mathcal{X}$ from a candidate pool of training data $\mathcal{D}_{pool}$ to form $ \mathcal{D}= \{ x_i  \}_{i=1}^N$. Secondly, we apply data augmentation to $\mathcal{D}$, obtaining $\hat{\mathcal{D}}=  \{  \{ a_{i}^{x_{1}}  \} _{i=1}^{M},  \{ a_{i}^{x_{2}}  \} _{i=1}^{M},…,  \{ a_{i}^{x_{N}}  \} _{i=1}^{M}  \}$, where $ \{ a_{i}^{x_{j}}  \} _{i=1}^{M}$ is the result of applying $M$ times augmentation on $x_{j}$. Next, we divide $ \hat{\mathcal{D}}$ into $K$ blocks $\hat{\mathcal{D}}=\left \{ \hat{\mathcal{D}}_1,...,\hat{\mathcal{D}}_K \right \}$, and each block contains $(N/K)\times M$ images. For humans, different images of the same object are easily categorized because of their high similarity. Similarly, as there is significant entity similarity in data augmented from the same image, we assign the same label $y_{j}\in\mathcal{Y}$ to the augmented data of the same image $x_{j}$ to simulate human classification thinking. We use $z=\{a,y\}\in \mathcal{Z}$ to represent the data that is categorized in the above way, where $y$ represents the label of the augmented sample $a$ of $x$, i.e., the dataset $\hat{\mathcal{D}}= \{ z_{i}  \}_{i=1}^{N\times M} $. Based on the analysis in Section\ref{Study of the Relationship}, each block in $\hat{\mathcal{D}}_i$ can be regarded as a $N/K$-way task $\mathcal{T} _{i}$, where $\hat{\mathcal{D}}= \{ \mathcal{T} _{i}  \}_{i=1}^{K} $. Take $\hat{\mathcal{D}}_1=\{ z_{i}  \}_{i=1}^{(N/K)\times M}$ as an example, we divide $\hat{\mathcal{D}}_1$ into two parts for the support set $\hat{\mathcal{D}}_1^{s}$ and query set $\hat{\mathcal{D}}_1^{q}$, respectively, each containing $(N/K)\times M_{1}$ and $(N/K)\times M_{2}$ images ($M_{1}+ M_{2}=M$). This approach transforms a batch of unlabeled data into $K$ $N/K$-way-$M_{1}$-shot classification tasks which realizes the advantages i and ii of human learning. 

\subsection{Bootstrapped Meta Self-supervised Learning}
\label{Bootstrapped Meta Self-supervised Learning}
We now describe how to use constructed few-shot tasks for meta-learning. As humans, we learn specific knowledge and build mind maps by going through two levels of abstraction, allowing us to quickly recognize similar things. Therefore, we focus on gradient-based methods \cite{finn2017model} that use two-level loops to constrain updates for task-specific and meta-learners to obtain optimal initial parameters $\theta$, which is analogous to human experience for faster adaptation to new tasks.

\textbf{For the inner loop (first optimization)}, the learner's objective is to learn $f(w)$ by minimizing the learning objective $ \left [ l(f(w);\theta,\mathcal{D} ) \right ] $ over the training data $\mathcal{D}$, where $\mathcal{D}$ represents a mini-batch of unlabeled data source, $f$ represents a neural network that consists of the feature extractor, projection head (contrastive learning), and classifier (cross entropy). We denote the parameter of $f(w)$ as $w$. Based on Subsection \ref{Online Task Construction}, $\mathcal{\mathcal{D}}$ can be extended to $K$ $N/K$-way-$M_{1}$-shot tasks. Considering the impact of label generation errors, the task-specific loss function consists of the cross-entropy loss $l_{ce}(\cdot)$ and the contrastive loss $l_{cl}(\cdot)$, expressed as:
\begin{equation}
\label{equ:1}
\begin{array}{l}
    l(f(w);\theta,\mathcal{D} )=l_{ce}(f(w);\theta,\mathcal{D})+\lambda l_{cl}(f(w);\theta;\mathcal{D})\\
    s.t.\quad l_{ce}(\cdot)=-\sum\nolimits_{j = 1}^{j = N} {\sum\nolimits_{i = 1}^{i = M} {{y_j}\log {\mathop a\limits^ \cdot}_i^{{x_j}}} } \\
    l_{cl}(\cdot)=- \sum\nolimits_{j = 1,i = 1}^{j = N,i = M} {\log \frac{{\sum\nolimits_{r = 1,r \ne i}^{r = M} {\exp (sim({\mathop a\limits^ *}_i^{{x_j}},{\mathop a\limits^ *}_r^{{x_j}})/\tau )} }}{{\sum\nolimits_{r = 1,r \ne i}^{r = M} {\exp (sim({\mathop a\limits^ *}_i^{{x_j}},{\mathop a\limits^ *}_r^{{x_j}}))}  + \sum\nolimits_{p = 1,p \ne j}^N {\sum\nolimits_{o = 1}^{o = M} {\exp (sim({\mathop a\limits^ *}_i^{{x_j}},{\mathop a\limits^ *}_p^{{x_o}})/\tau )} } }}} 
\end{array}
\end{equation}
where $\lambda $ describes the importance of $l_{cl}(\cdot)$, ${\mathop a\limits^ \cdot}_i^{{x_j}}$ is the output of classifier, and ${\mathop a\limits^ *}_i^{{x_j}}$ is the output of projection head. Finally, based on an initial parameter $\theta^0$ of $w$, we train a task-specific model with learning rate $\alpha$ and obtain the weights $w=\theta^0-\alpha \bigtriangledown _{w}l(w;\theta^0,\mathcal{D} )$.

\textbf{For the outer loop (second optimization)}, the aim of meta-learner is to learn the model $F$, e.g., $F(\mathcal{D} )=f(w)$, which means generating the optimal task-specific model $f(w)$ given dataset $\mathcal{D}$. In machine learning, $f$ needs to be obtained through a series of gradient descent updates based on the loss function $l(\cdot)$, starting from an initial parameter, and it is difficult to reach the optimal result in a single step. Therefore, the objective of the meta-learner is transformed into finding a learnable optimal initialization $\theta \in \zeta$ to let $f(w)$ to quickly obtain the optimal parameter $w \in \mathcal{W}$. Thus, meta-learning formulate the objective function of learning $F$ as ${\min _\theta }l(f({w^*}(\theta ));\theta ,\mathcal{D}),s.t.,{w^*}(\theta ) = \arg {\min _w}l(f(w);\theta ,\mathcal{D})$. We can see that the standard meta-gradient firstly optimizes constraint condition ${w^*}(\theta ) = \arg {\min _w}l(f(w);\theta ,\mathcal{D})$ by performing $L$ steps updates  and then evaluating ${w^*}(\theta )$ under $l(f({w^*}(\theta ));\theta ,\mathcal{D})$, thus obtaining the update rule of $\theta$:
\begin{equation}
\label{equ:2}
    \theta^{'}=\theta -\beta \bigtriangledown_{\theta}l(f({w^*}(\theta ));\theta ,\mathcal{D})
\end{equation}
However, although this method can introduce prior experience $\theta$ like humans and quickly adapt to new tasks, the learning process of $\theta$ still has limitations: i) it highly depends on $f(w)$, while human thinking is unrestricted; ii) it is based on updates within a limited number of steps $L$, while human learning is extendable. Therefore, considering that human induction is based on entity similarity, we convert this experiential learning into a metric in the model: using the bootstrapped target to move ${w^{L}}(\theta )$ closer to its $L+\delta$ version ${w^{L+\delta}}(\theta)$. We regard ${w^{L}}(\theta )$ and ${w^{L+\delta}}(\theta)$ as two discrete uniform distributions with respect to their constituent units. To expand the perspective of updating $\theta$, for the outer loop, we use KL divergence to bring the distribution $\pi_{w^{(L)}}$ obtained by ${w^{L}}(\theta )$ step closer to the distribution $\pi_{\vec{w}}$ obtained by ${w^{L+\delta}}(\theta)$ to bootstrapped state ${w^{L}}(\theta )$ extended to  ${w^{L+\delta}}(\theta)$, thereby encouraging the meta-learner to reach future states on its trajectory faster. We have:
\begin{equation}
\label{equ:3}
\begin{array}{l}
    \vec{\theta}=\theta-\beta \bigtriangledown_{\theta} D_{KL} (\pi_{\vec{w}},\pi_{w^{L}})\\
\end{array}
\end{equation}
where ${w^{L}}(\theta )$ is to update $L$ step based on ${w^*}(\theta ) = \arg {\min _w}l(f(w);\theta ,\mathcal{D})$, while ${w^{L+\delta}}(\theta)$ is to update $L+\delta$ step.

This measure eliminates the dependence of the outer update $\theta$ on the task-specific model $f$, and makes $\pi_{w^{L}}$ continuously approach $\pi_{\vec{w}}$ that contains future information to achieve convergence and break through the limitation of limited updates. This paradigm makes the model its own teacher.

\subsection{Theoretical Analysis}
\label{Theoretical Analysis}
We now conduct a theoretical analysis of our BMSSL for performance guarantees. We defer all proofs and more analysis to Appendix A.

First, consider task construction described in Subsection \ref{Online Task Construction}, the goal is to identify a representation that allows us to approximate many different "reasonable" choices by $g$. It can group augmented views of similar entities together, i.e. every $g$ that satisfies the following assumption:


\begin{assumption}\label{assum:AVI}
    \rm{(Approximate View-Invariance)}: \emph{The best estimate of the label $y$ is approximately invariant to the choice of different augmented views $a$ of the same $x$. Each target function $g:\mathcal{A} \to \mathbb{R}^n $ satisfies:
    \begin{equation}
    \mathbb{E}_{p_+(a_1,a_2)}\left [ (g(a_1)-g(a_2))^2 \right ]\le \varepsilon   
    \end{equation}
    where $p_+(a_1,a_2)=\sum_{x}p(a_1|x)p(a_2|x)p(x)$ when fixed $\varepsilon \in [0,\infty)$.
    }
\end{assumption}

We can then constrain the error of approximating $g$ with a small subset of eigenfunctions by constraining each coefficient according to its contribution to the total positive pair difference. We focus on a class of linear predictors on top of $k$-dimensional representations $r:\mathcal{A}\to \mathbb{R}^n  $, among which the representation $r^d=\left \{ p_1(a),p_{2}(a),...,p_d(a) \right \} $ contains $d$ eigenfunctions of the positive-pair Markov chain with the largest eigenvalues is the best choice and used for task-specific training (Equation \ref{equ:1}).

\begin{theorem}\label{theorem:inner}
    \rm{(Task-specific Performance Guarantee)} \emph{Let $\mathcal{G}_\varepsilon $ be the functions satisfying Assumption \ref{assum:AVI}, and $\mathcal{G}_r=\left \{ a\mapsto \hat{g}_\nu (a)= \nu^T r(a)\right \} $ be the subspace of linear predictors which maximizes the view invariance of the least-invariant unit-norm predictor $\mathcal{G}_{r^d}$ with implicit regularization effect:
    \begin{equation}
        \mathcal{G}_{r^d}=\mathop{\arg\min}_{\mathcal{G}}\mathop{\max}_{{\hat{g}\in\mathcal{G}}}\mathbb{E}_{p_+(a_1,a_2)}\left [ (\hat{g}(a_1)-\hat{g}(a_2))^2 \right ],\quad \text{s.t. }\text{dim}(\mathcal{G})=d,\ \mathbb{E}[\hat{g}(a)^2=1]
    \end{equation}
    it is implied in $\mathcal{G}$, which minimizes the approximation error of the worst-case objective function $f$:
    \begin{equation}
        \mathcal{G}_{r^d}=\mathop{\arg\min}_{\mathcal{G}}\mathop{\max}_{g\in \mathcal{G}_\varepsilon}\mathop{\min}_{{\hat{g}\in\mathcal{G}}}\mathbb{E}_{p_+(a_1,a_2)}\left [ (\hat{g}(a_1)-\hat{g}(a_2))^2 \right ]
    \end{equation}
    }
\end{theorem}

Theorem \ref{theorem:inner} states that the function class we built (Equation \ref{equ:1}) is the best choice for the least squares approximation that satisfies Assumption \ref{assum:AVI}. Next, we turn to bootstrapped meta-training in outer loop, where we take Assumption \ref{assum:Task}-\ref{assum:OF} mentioned in Section \ref{Study of the Relationship}.

\begin{theorem}
\textbf{(Bootstrapped Meta-training Performance Guarantee)} \emph{Let $w$ and $\theta$ be given by Equations mentioned in Section \ref{Bootstrapped Meta Self-supervised Learning}, the update process satisfies:
\begin{equation}
    f(w^L(\vec{\theta}))-f(w^L(\theta ))=\frac{\beta }{\alpha }(D_{KL}(\vec{w},w^L-\alpha \nabla_{w}f(w^L))-D_{KL}(\vec{w},w^L))+o(\beta(\alpha+\beta) ) 
\end{equation}
Let $\vec{\theta}$ and $\theta'$ be given by Equation \eqref{equ:3} and \eqref{equ:2} respectively, $f(w^L(\vec{\theta}))-f(w^L(\theta ))\le 0$ when $(\alpha, \beta)$ sufficiently small, while $f(w^L(\theta'))-f(w^L(\theta ))\le 0$ when $\beta$ sufficiently small. With the state $\vec{w}$ bootstrapped from $w^L$ with $\delta$ steps which offer future distribution (better), the update process will turn into:
\begin{equation}
    f(w^L(\vec{\theta}))-f(w^L(\theta ))=-\frac{\beta }{\alpha }D_{KL}(\vec{w},w^L)+o(\beta(\alpha+\beta) )<0 
\end{equation}
}
\end{theorem}

Therefore, compared to standard meta-learning, BMSSL enables models to reach optimal results faster while achieving convergence without utilizing gradient updates.


\section{Empirical Study}
\label{Experiments}
In this section, we conduct several experiments to benchmark and analyze BMSSL, including standard self-supervised few-shot learning (Subsection \ref{Standard}), cross-domain self-supervised few-shot learning (Subsection \ref{Cross-domain}), and ablation study (Subsection \ref{Ablation Studies}). The details of implementation are available at Appendix B. We omit the confidence intervals in this section for clarity, and the full results with them are provided in Appendix E. Our main objective is to demonstrate the effectiveness of BMSSL by exploring two key questions: i) Can BMSSL be applied successfully in self-supervised few-shot classification scenarios and achieve superior generalization performance? ii) By simulating the way humans learn, can BMSSL achieve more robust learning outcomes?


\subsection{Standard Self-supervised Few-shot Learning}
\label{Standard}
\textbf{Setup}. We evaluate BMSSL on three standard few-shot benchmarks of unsupervised meta-learning: Omniglot \cite{lake2019omniglot}, \emph{mini}ImageNet \cite{vinyals2016matching}, and CIFAR-FS \cite{bertinetto2018meta}. Following \cite{jang2023unsupervised}, we compare the performance of BMSSL with unsupervised meta-learning methods \cite{hsu2018unsupervised, khodadadeh2019unsupervised, khodadadeh2020unsupervised, lee2021meta, kong2021unsupervised, jang2023unsupervised}, self-supervised learning methods \cite{chen2020simple, zbontar2021barlow, he2020momentum, caron2020unsupervised}, and supervised meta-learning methods \cite{finn2017model, snell2017prototypical, requeima2019fast}. To explore the effect of simulating the way humans learn, we introduce the standard meta self-supervised learning paradigm (MetaSSL) mentioned in Section \ref{Algorithm}, which is also a two-layer structure but not optimized by bootstrapped target. See Appendix C and D for details on benchmarks and baselines.

\textbf{Results}. Table \ref{tab:standard} presents the results of few-shot classification on various (way, shot) tasks for the three benchmark datasets mentioned above. We obtain the following three observations: i) outstanding performance: BMSSL achieves outstanding performance on all three benchmarks, surpassing previous unsupervised meta-learning SOTA models. For instance, we obtain an accuracy gain of 3.763\% in the 20-way-1-shot test. ii) improve generalization: its performance is competitive even in the unsupervised setting compare to supervised meta-learning and self-supervised baselines. iii) effectiveness of simulate human learning: we compare our approach with standard MetaSSL and achieve an average gain of 4.045\%.

\begin{table}
  \caption{Accuracy (\%) of standard few-shot classification on Omniglot, \emph{mini}ImageNet, and CIFAR-FS benchmarks. The values in this table are average accuracies over 2000 "\textbf{(way, shot)}" tasks for BMSSL and baselines following \cite{jang2023unsupervised}. \textbf{Bold} entries indicate the best for unsupervised tasks. \colorbox{blue!10}{Blue} indicates our BMSSL before and after introducing bootstrapped mentioned in Section \ref{Bootstrapped Meta Self-supervised Learning}.}
  \label{tab:standard}
  \centering
\resizebox{\linewidth}{!}{\begin{tabular}{lccccccccc}
  \toprule
\multirow{2}{*}{\textbf{Method}} & \multicolumn{3}{c}{\textbf{Omniglot}} & \multicolumn{3}{c}{\textbf{\emph{mini}ImageNet}} & \multicolumn{3}{c}{\textbf{CIFAR-FS}}\\
  \cmidrule(r){2-4}
  \cmidrule(r){5-7}
  \cmidrule(r){8-10}
    & \textbf{(5,1)} & \textbf{(5,5)} & \textbf{(20,1)} & \textbf{(5,1)} & \textbf{(5,5)} & \textbf{(20,1)} & \textbf{(5,1)} & \textbf{(5,5)} & \textbf{(20,1)}\\
    \midrule
    \emph{Train from scratch} & 50.29 & 72.82 & 26.20 & 24.20 & 38.84 & 16.29 & 31.12 & 44.89 & 20.32  \\
    \midrule
    \rowcolor{gray!40}\multicolumn{10}{c}{\emph{Unsupervised Meta-learning}}\\
    \midrule
    CACTUs\cite{hsu2018unsupervised} & 65.29 & 86.25 & 49.54 & 39.32 & 53.54 & 31.99 & 40.02 & 58.16 & 35.88  \\
    UMTRA\cite{khodadadeh2019unsupervised}& 83.32 & 94.23 & 75.84 & 39.23 & 51.78 & 30.27 & 41.61 & 60.55 & 37.10  \\
    LASIUM\cite{khodadadeh2020unsupervised}& 82.38 & 95.11 & 70.23 & 42.12 & 54.98 & 34.26 & 45.33 & 62.65 & 38.40  \\
    Meta-SVEBM\cite{kong2021unsupervised}& 87.07 & 94.13 & 73.33 & 44.74 & 58.38 & 39.71 & 47.24 & 63.10 & 40.10  \\
    Meta-GMVAE\cite{lee2021meta}& 90.89 & 96.05 & 81.51 & 42.28 & 56.97 & 39.83 & 47.45 & 63.20 &  41.55 \\
    PsCo\cite{jang2023unsupervised}& \textbf{96.18} & 98.22 & 89.32 & 46.35 & 63.05 & 40.84 & 51.77 & \textbf{69.66} & 45.08 \\
    \rowcolor{blue!10}MetaSSL& 91.36 & 95.35 & 88.64 & 45.82 & 62.14 & 39.48 & 49.09 & 66.54 & 43.70  \\
    \rowcolor{blue!10}BMSSL & 96.02 & \textbf{99.56} & \textbf{91.41} & \textbf{49.98} & \textbf{64.59} & \textbf{45.28} & \textbf{52.20} & 69.64 & \textbf{49.84}  \\
    \midrule
    \rowcolor{gray!40}\multicolumn{10}{c}{\emph{Self-supervised Learning}}\\
    SimCLR\cite{chen2020simple} & 90.83 & 97.67 & 81.67 & 42.32 & 51.10 & 36.36 & 49.44 & 60.02 & 39.29  \\
    MoCo\cite{he2020momentum}& 87.83 & 95.52 & 80.03 & 40.56 & 49.41 & 36.52 & 45.35 & 58.11 & 37.89  \\
    SwAV\cite{caron2020unsupervised}& 91.28 & 97.21 & 82.02 & 44.39 & 54.91 & 37.13 & 49.39 & 62.20 & 40.19  \\
    \midrule
    \rowcolor{gray!40}\multicolumn{10}{c}{\emph{Supervised Meta-learning}}\\
    MAML\cite{finn2017model} & 93.22 & 97.53 & 82.36 & 45.84 & 63.25 & 36.77 & 48.25 & 58.00 & 39.52  \\
    ProtoNet\cite{snell2017prototypical}& 95.83 & 99.29 & 92.80 & 46.58 & 63.20 & 40.11 & 51.28 & 69.55 & 46.65 \\
    CNAPs\cite{requeima2019fast}& 91.28 & 95.98 & 87.09 & 43.21 & 62.87 & 36.55 & 52.07 & 70.38 & 43.30 \\
    \bottomrule
  \end{tabular}}
\end{table}

\subsection{Cross-domain Self-supervised Few-shot Learning}
\label{Cross-domain}
\textbf{Setup}. We compare the effect of BMSSL and the baseline described by Seciton \ref {Standard} on the cross-domain few-shot classification benchmarks, which is divided into two categories based on the similarity with ImageNet: i) high similarity: CUB \cite{welinder2010caltech}, Cars \cite{krause20133d}, and Places \cite{zhou2017places}; ii) low similarity: CropDiseases \cite{mohanty2016using}, ISIC \cite{codella2018skin}, and ChestX \cite{wang2017chestx}.

\textbf{Results}. Table \ref{tab:cross-domain} presents the performance of the model trained on \emph{mini}ImageNet for meta-learning on the benchmark datasets mentioned above. By observation, we further validate the performance of our proposed BMSSL: i) Effectiveness: achieves similar or even better results than the state-of-the-art baseline algorithms on all benchmark datasets; ii) Generalization: achieves nearly a 3\% improvement compared to supervised meta-learning and self-supervised learning on the datasets with significant differences from the training phase; iii) Robustness: achieves similar results to the PsCo \cite{jang2023unsupervised} which introduces out-of-distribution samples, even though we do not explicitly consider out-of-distribution samples on datasets with significant differences.
\begin{table}
  \caption{Accuracy (\%) of cross-domain few-shot classification with two types mentioned in Section \ref{Cross-domain}. We transfer models trained on \emph{mini}ImageNet to each benchmark. The meanings of "\textbf{(way, shot)}", "\textbf{Bold}" and "\colorbox{blue!10}{Blue}" in the table are consistent with Table \ref{tab:standard}.}
  \label{tab:cross-domain}
  \centering
\resizebox{\linewidth}{!}{\begin{tabular}{lcccccccccccc}
  \toprule
\multirow{2}{*}{\textbf{Method}} & \multicolumn{2}{c}{\textbf{CUB}} & \multicolumn{2}{c}{\textbf{Cars}} & \multicolumn{2}{c}{\textbf{Places}} & \multicolumn{2}{c}{\textbf{CropDiseases}} & \multicolumn{2}{c}{\textbf{ISIC}} & \multicolumn{2}{c}{\textbf{ChestX}}\\
  \cmidrule(r){2-3}
  \cmidrule(r){4-5}
  \cmidrule(r){6-7}
  \cmidrule(r){8-9}
  \cmidrule(r){10-11}
  \cmidrule(r){12-13}
    & \textbf{(5,5)} & \textbf{(5,20)} & \textbf{(5,5)} & \textbf{(5,20)} & \textbf{(5,5)} & \textbf{(5,20)} & \textbf{(5,5)} & \textbf{(5,20)} & \textbf{(5,5)} & \textbf{(5,20)} & \textbf{(5,5)} & \textbf{(5,20)}\\
    \midrule
    \rowcolor{gray!40}\multicolumn{13}{c}{\emph{Unsupervised meta-learning}}\\
    \midrule
    Meta-SVEBM  & 45.893 & 54.823 & 33.530 & 44.622 & 50.516 & 61.561 & 71.652 & 84.515 & 37.106 & 48.001 & 27.238 & 29.652 \\
    Meta-GMVAE & 48.783 & 55.651 & 30.205 & 39.946 & 55.361 & 65.520 & 72.683 & 80.777 & 30.630 & 37.574 & 24.522 & 26.239 \\
    PsCo & 56.365 & 69.298 & 44.632 & 56.990 & 64.501 & 73.516 & \textbf{89.565} & 95.492 & 43.632 & 54.886 & 21.907 & 24.182 \\
    \rowcolor{blue!10}MetaSSL & 54.238 & 65.031 & 45.341 & 56.526 & 62.538 & 70.022 & 83.922 & 90.058 & 40.140 & 50.209 & 24.827 & 25.238 \\
    \rowcolor{blue!10}BMSSL  & \textbf{57.543} & \textbf{69.561} & \textbf{49.636} & \textbf{59.511} & \textbf{67.250} & \textbf{75.834} & 87.524 & \textbf{95.950} & \textbf{46.518} & \textbf{56.293} & \textbf{29.463} & 30.389 \\
    \midrule
    \rowcolor{gray!40}\multicolumn{13}{c}{\emph{Self-supervised Learning}}\\
    SimCLR  & 51.389 & 60.011 & 38.639 & 52.412 & 59.523 & 68.419 & 80.360 & 89.161 & 44.669 & 51.823 & 26.556 & \textbf{30.982} \\
    MoCo & 52.843 & 61.204 & 39.504 & 50.108 & 60.291 & 69.033 & 81.606 & 90.366 & 44.328 & 52.398 & 24.198 & 27.893 \\
    SwAV & 51.250 & 61.645 & 36.352 & 51.153 & 58.789 & 68.512 & 80.055 & 89.917 & 43.200 & 50.109 & 21.252 & 28.270 \\
    \midrule
    \rowcolor{gray!40}\multicolumn{13}{c}{\emph{Supervised Meta-learning}}\\
    MAML  & 57.296 & 64.005 & 44.934 & 49.561 & 62.502 & 71.741 & 78.202 & 85.247 & 46.405 & \textbf{56.293} & 22.435 & 24.238 \\
    ProtoNet & 56.237 & 64.829 & 40.893 & 48.123 & 59.887 & 69.207 & 76.651 & 84.164 & 40.028 & 49.289 & 22.219 & 25.839 \\
    \bottomrule
  \end{tabular}}
\end{table}


\subsection{Ablation Studies}
\label{Ablation Studies}

\textbf{Augmentations in task construction}. Although we have shown in Figure \ref{fig:intro_fig} that augmentation cannot offset the impact of data scarcity, we have not yet explored the effects of different levels of augmentation on SSL task construction, which is directly related to the diversity and feature similarity of the samples in the task. We divide augmentation methods into four levels with different quantities (five kinds/one kind) and intensities (mild/strong, such as large/small area splicing), and apply them to evaluate the impact on the model. The experimental results in Table \ref{tab:ablation1} shows that the benefits of data diversity to the model are limited, and augmentation strategies have little effect on the model.

\textbf{The effect of bi-level optimization}. BMSSL introduces learning experience through two loops of gradient update and gives the model the ability to optimize and constrain twice. To evaluate the its effect, we fix the constraint structure of the inner loop and compared it under the following three settings: i) one-time optimization + no introduction of experience ($\mathcal{M}_1 $): only containing task-specific models with random initialization; ii) one-time optimization + introduction of experience ($\mathcal{M}_2$): only updating the model's metric learning for specific tasks, and using it as experience; iii) two-time optimization + introduction of experience ($\mathcal{M}_3$): BMSSL's bi-level optimization structure. The results of this ablation experiment are shown in Table \ref{tab:ablation2}. BMSSL achieves nearly a 4\% improvement, demonstrating the gain from priors and model structure on the algorithm.

\textbf{Training $L$ and $\delta$}. To find the optimal parameters of the model, we test the model on \emph{mini}ImageNet with different settings of $L$ and $\delta$. Tables \ref{tab:ablation3} shows the model's accuracy and running efficiency when $L=5$, which runs on a V100 NVIDIA GPU. We find $\delta=5$ may be the best choice where the further increase has little effect on the accuracy, but the operating efficiency drops greatly. The introduction of $\delta$ can adjust the efficiency of the model and achieve faster convergence through this distribution approximation. The full results and further analysis areavailable at Appendix E.

\begin{minipage}{\textwidth}
\begin{minipage}[t]{0.3\textwidth}
\makeatletter\def\@captype{table}
\caption{Accuracy(\%) on \emph{mini}ImageNet with four levels of data augmentation expressed as $\left \{ \mathcal{A}_i \right \}_{i=1}^4 $.}
\begin{tabular}{cc}
    \toprule
    \textbf{Levels} & \textbf{Top-1 Acc(\%)} \\
    \midrule
    $\mathcal{A}_1$  & 49.730$\pm$0.303 \\
    $\mathcal{A}_2$  & 49.990$\pm$0.238 \\
    $\mathcal{A}_3$  & 49.789$\pm$0.210 \\
    $\mathcal{A}_4$  & 49.832$\pm$0.199 \\
    \bottomrule
\end{tabular}
\label{tab:ablation1}
\end{minipage}
\hspace{.05in}
\begin{minipage}[t]{0.3\textwidth}
\makeatletter\def\@captype{table}
\caption{Accuracy(\%) on \emph{mini}ImageNet under three types of model structures mentioned in Section \ref{Ablation Studies}.}
\begin{tabular}{cc}
    \toprule
    \textbf{Method} & \textbf{Top-1 Acc(\%)} \\
    \midrule
    $\mathcal{M}_1$  & 39.583$\pm$0.482 \\
    $\mathcal{M}_2$  & 46.184$\pm$0.298 \\
    $\mathcal{M}_3$  & 49.987$\pm$0.283 \\
    \bottomrule
\end{tabular}
\label{tab:ablation2}
\end{minipage}
\hspace{.05in}
\begin{minipage}[t]{0.37\textwidth}
\makeatletter\def\@captype{table}
\caption{Accuracy(\%) and meta-training steps($/s$) when $L=5$ on \emph{mini}ImageNet with different $\delta$.}
\begin{tabular}{ccc}
    \toprule
    \textbf{$\delta$} & \textbf{Top-5 ACC(\%)}  & \textbf{Steps($/s$)} \\
    \midrule
    1        & 63.832    & 4.3  \\
    5        & 64.443    & 3.2  \\
    10          & 64.592    & 2.6  \\
    15          & 64.588    & 2.1  \\
    20          & 64.605    & 1.7  \\
    \bottomrule
\end{tabular}
\label{tab:ablation3}
\end{minipage}
\end{minipage}





\section{Conclusion}
\label{Conclusion}
In this work, we propose a novel Bootstrapped Meta Self-Supervised Learning framework that simulates three advantages of human learning: i) without supervision, ii) not limited by data, and iii) learning from experience. We discuss the relationship between self-supervised learning and meta-learning, and leveraging our findings to propose a simple but clever approach for reconstructuring self-supervised tasks. Additionally, we employ bi-level optimizations to introduce experience for learning, and use meta-gradients to generate bootstrapped target to make the model its own teacher. Through theoretical analysis and extensive experiments, we demonstrate the superior performance of our framework.

\textbf{Broader Impact and Limitations}. This work offers a reliable way for machines to mimic human learning, providing technological advances in machine learning. We do not need to be data-bound or long-term training as previous methods. But it has a limitation that the evaluation focuses on visual tasks, without considering the learning effects of other fields (e.g., reinforcement learning and language recognition) or other tasks (e.g., regression, generation). 


\bibliography{neurips2023}{}
\bibliographystyle{plain}

\end{document}